\documentclass{article}

\PassOptionsToPackage{numbers, compress}{natbib}

\usepackage[preprint]{neurips_2021}

\usepackage[utf8]{inputenc} 
\usepackage[T1]{fontenc}    
\usepackage{hyperref}       
\usepackage{url}            
\usepackage{booktabs}       
\usepackage{amsfonts}       
\usepackage{nicefrac}       
\usepackage{microtype}      
\usepackage{xcolor}         

\usepackage{graphicx}
\usepackage{epsfig}
\usepackage{amsmath}
\usepackage{amssymb}
\usepackage{caption}
\usepackage{subcaption}
\usepackage{multirow}
\usepackage{amssymb}
\usepackage{pifont}
\newcommand{\cmark}{\ding{51}}%
\newcommand{\xmark}{\ding{55}}%
\usepackage{enumitem}
\usepackage[ruled,vlined]{algorithm2e}
\usepackage{authblk}
\usepackage{wrapfig}
\usepackage{floatrow}
\newfloatcommand{capbtabbox}{table}[][\FBwidth]

\newcommand{\modelnameshort}{FNAS}
\newcommand{\hs}[1]{{\color{black} #1}}

\title{FNAS: Uncertainty-Aware Fast Neural Architecture Search}

\author[1, 2]{Jihao Liu}
\author[1]{Ming Zhang}
\author[1]{Yangting Sun}
\author[1]{Boxiao Liu}
\author[1, 2]{Yu Liu}
\author[2, 3]{Hongsheng Li}

\affil[1]{\footnotesize SenseTime X-Lab}
\affil[2]{\footnotesize CUHK - SenseTime Joint Lab, The Chinese University of Hong Kong}
\affil[3]{\footnotesize School of CST, Xidian University}

\begin{document}

\maketitle

\begin{abstract}
  Reinforcement learning (RL)-based neural architecture search (NAS) \hs{generally} guarantees better convergence yet suffers from \hs{the requirement of} huge computational resources compared with gradient-based approaches, due to the \textit{rollout bottleneck} -- exhaustive training of each sampled architecture on the proxy tasks. 
  \hs{In this paper, we propose} a general pipeline to accelerate the convergence of the rollout process as well as the RL process in NAS. It is motivated by the interesting observation that both the architecture and the parameter knowledge can be transferred between different search processes and even different tasks.
  We first introduce an uncertainty-aware critic (value function) in Proximal Policy Optimization (PPO) \cite{ppo} to take advantage of the architecture knowledge in previous search processes, which stabilizes the training process and reduce the searching time by 4 times.
  In addition, an architecture knowledge pool together with a block similarity function is proposed to utilize parameter knowledge and reduces the searching time by 2 times. To the best of our knowledge, this is the first method that introduces a block-level weight sharing scheme in RL-based NAS. The block similarity function guarantees a 100\% hit ratio with strict fairness \cite{fairnas}. 
  Besides, we show \hs{that} an off-policy correction factor used in ``replay buffer'' of RL optimization can further reduce half of the searching time.
  Experiments on the Mobile Neural Architecture Search (MNAS) \cite{MnasNet} search space show that the proposed Fast Neural Architecture Search (\modelnameshort{}) accelerates the standard RL-based NAS process by $\sim$10x (e.g., 20,000 GPU hours to 2,000 GPU hours for MNAS), and guarantees better performance on various vision tasks.
\end{abstract}

\section{Introduction}

The architecture of a convolutional neural network (CNN) is crucial for many deep learning tasks such as image classification \cite{efficientnet} and object detection \cite{tan2019efficientdet}. The widespread use of  neural architecture search (NAS) methods such as differentiable, one-shot,
evolutional, and RL-based approaches have  effectively dealt with architecture design problems.
Despite having high performance due to its sampling-based mechanism \cite{MnasNet, NASNet, efficientnet}, RL-based NAS tends to require unbearable computing resources which discourages the research community from exploring it further.

The main obstacles to the propagation of RL-based NAS algorithm come from the following two aspects: a) it's necessary to sample a large number of architectures from the search space to ensure the convergence of the RL agent, b) the inevitable training and evaluation cost of these \hs{architecture} samples on proxy tasks. For example, the seminal RL-based NAS \cite{NAS-RL} approach requires 12,800 generations of architectures. The state-of-the-art MNAS \cite{MnasNet} and MobileNet-V3 \cite{mbv3} require 8000 or more generations to \hs{find} the optimal architecture. 
Coupled with $\sim$5 epochs training for each generation, the whole search process costs nearly 64 TPUv2 devices for 96 hours or 20,000 GPU hours on V100 for just one single searching process. With no access to reduce the unbearable computational cost, RL-based NAS is hard to make more widespread influence than differential \cite{DARTS, PDARTS}, and one-shot based \cite{bender2019understanding,guo2019single} methods.

On the contrary, the high efficiency of one-shot NAS family brings it continuous research attention. Instead of sampling a huge number of sub-networks, one-shot NAS assembles them into a single super-network. The parameters are shared between different sub-networks during the training of the super-network. In this way, the training process is condensed from training thousands of sub-networks into training a super-network. However, this \hs{weight sharing} strategy may \hs{cause} problems \hs{of inaccurate} performance estimation of sub-networks. For example, two sub-networks may propagate conflicting gradients to their shared components, \hs{which} may converge to favor one of the sub-networks and repel the other \hs{one} randomly. This conflicting phenomenon may result in instability of the search process and inferior final architectures, compared with RL-based methods.

In this work, we \hs{aim at combining advantages} of both RL-based methods and one-shot methods.
The proposed method is based on two \hs{important \textit{key observations}}:
First, the optimal architectures for different tasks have \hs{certain} common architecture knowledge (similar sub-architectures in different search processes' optimal architectures). 
Second, the parameter knowledge (weights at samples' training checkpoints) can also be transferred across different searching settings and even tasks. 

Based on the two observations, to transfer architecture knowledge, we develop Uncertainty-Aware Critic (UAC) to learn the architecture-performance joint distribution from previous search processes in an unbiased manner, utilizing the transferability of the architecture knowledge, which reduces the needed samples in RL optimization process by 50\%.
For the transferable \textit{parameter knowledge}, we propose an Architecture Knowledge Pool (AKP) to restore the block-level \cite{MnasNet} parameters and fairly share them as new sample architectures' initialization, which speed up each sample's convergence for $\sim$2 times.
Finally, we also develop an Architecture Experience Buffer (AEB) with an off-policy correctness factor to store the previously trained models for reusing in RL optimization, with half of the search time saved. 
Under the same environment as MNAS \cite{MnasNet} with MobileNet-v3 \cite{mbv3}, FNAS speeds up the search process by 10$\times$ and the searched architecture performs even better.

To summarize, our main contributions are as follows:
\begin{enumerate}
  \item We propose FNAS, which introduces three acceleration modules, uncertainty-aware critic, architecture knowledge pool, and architecture experience buffer, to speed up reinforcement-learning-based neural architecture search by $\sim$10$\times$.
  
  \item We show that the knowledge of neural architecture search processes can be transferred, which is utilized to improve sample efficiency of reinforcement learning agent process and training efficiency of each sampled architecture.
  
  \item We demonstrate new state-of-the-art accuracy on ImageNet classification, face recognition, and COCO object detection with comparable computational constraints.
\end{enumerate}

\begin{figure*}[t!]
\centering
\centerline{\includegraphics[width=\textwidth]{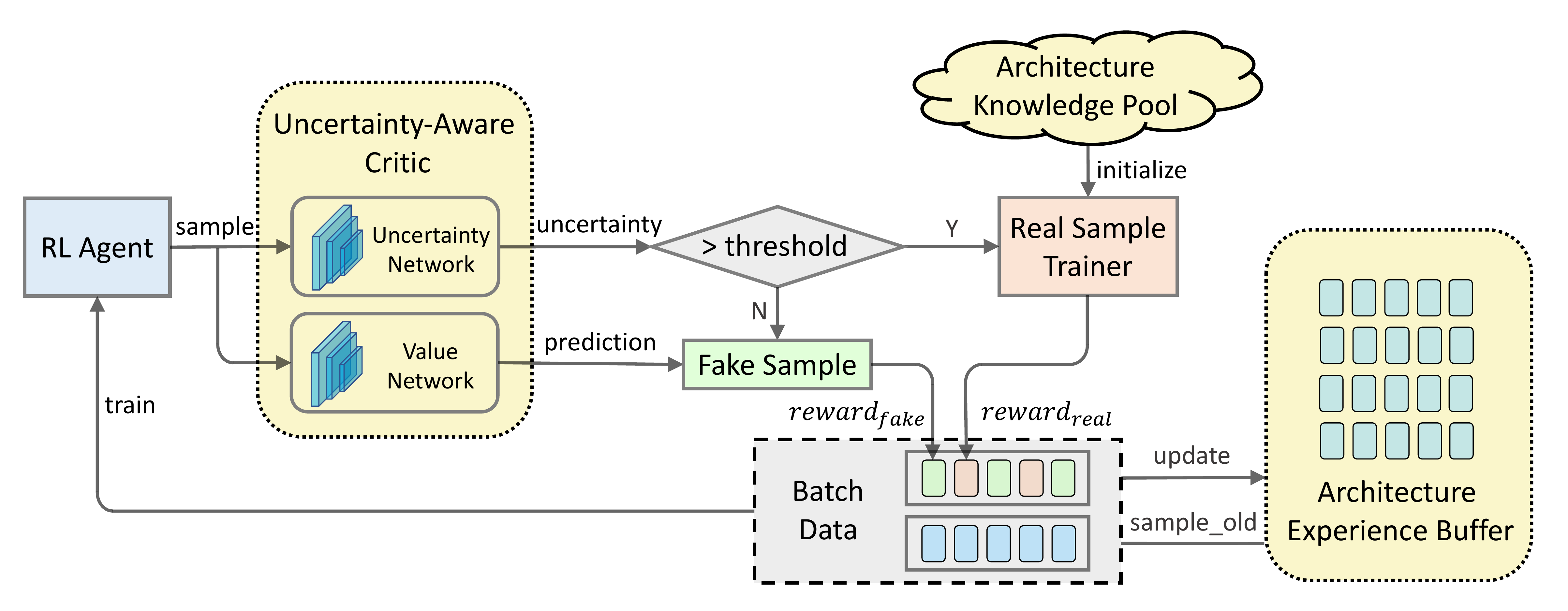}}
\caption{The pipeline of FNAS. The proposed modules are highlighted in orange. Architectures are sampled by the RL agent and then passed to Uncertainty-Aware Critic (UAC) for predicting performance and the corresponding uncertainty. Then a decision module will determine whether the sample needs to be trained by Trainer. The Architecture Knowledge Pool (AKP) helps to initialize new samples for training. Half of the samples in one batch come from Architecture Experience Buffer (AEB), the other half come from Trainer or UAC's Value Network.}
\label{fig.pipeline}
\end{figure*}

\section{Related Works}
\label{revisiting}

From the perspective of how to \hs{estimation the performance} of architectures, NAS methods can be \hs{classified} into two categories, sampling-based and \hs{weight-sharing}-based \hs{methods}. 

Sampling-based methods \hs{generally sample a large number of} architectures from the architecture search space and train them independently. Based on the \hs{evaluated} performance of the well-trained \hs{sampled} architectures, \hs{multiple} approaches can be utilized to \hs{identify} the \hs{best-performing} one, \hs{including} Bayesian optimization \cite{NAS-BO}, evolutionary algorithm \cite{AmoebaNet}, and \hs{optimization of an RL agent} \cite{NAS-RL}. The main drawback of \hs{this type of methods} is their tremendous time and computational consumption \hs{on} training the sampled architectures. To alleviate this issue, a common practice is to shorten the training epochs and use proxy networks with fewer filters and cells  \cite{NASNet, MnasNet}. Besides, Liu \emph{et al.} \cite{pnas} \hs{proposed} to train a network to predict the final performance. We also aim to reduce the training cost, by leveraging the accumulated architecture knowledge and parameter knowledge to accelerate the searching process.

Instead of training many architectures independently, the second \hs{type} of methods resort to training a super-network and estimate the performance of architectures with shared weights from the super-network \cite{bender2019understanding, wu2019fbnet, DARTS, PDARTS, xu2019pc, OnceForAll,stamoulis2019single, guo2019single}. With the easy access to performance estimation of each sub-architecture, DARTS \cite{DARTS} \hs{introduced} a gradient-based method to search for the best architecture in an end-to-end manner. However, as pointed in \cite{li2019random}, the estimated architecture performances based on \hs{weight-sharing networks might} be unreliable. Chen \emph{et al.} \cite{PDARTS} \hs{proposed} to progressively shrink the search space so that the estimation can be \hs{gradually} more accurate. Cai \emph{et al.} \cite{OnceForAll} \hs{introduced} a shrinking based method to train the \hs{super-network} so as to generate networks of different scales without re-training.

Besides, some existing works have tried to combine these two types of methods and reserve both of their advantages~\cite{shi2019bridging, zhao2020few}. BONAS, introduced in ~\cite{shi2019bridging}, is a sampling-based algorithm that utilizes weight-sharing to evaluate a batch of architectures simultaneously, which reduces the training cost significantly. Although weight-sharing in a batch can make the training more fair, the sub-networks in a batch are selected based on Bayesian Optimization method and can interfere each other in the training process, making the estimate of performance unreliable. Zhao \emph{et al.}~\cite{zhao2020few} propose to use multiple super-networks to alleviate the undesired co-adaption, which is highly sensitive to the splitting strategy of the search space. Cai \emph{et al.}~\cite{cai2018efficient} propose to transform the architecture repeatedly in the search process, where the weight of network can be reused to save computational cost. \hs{In our pipeline, we also propose to share weights between architectures but in a different way}. We construct a general weight pool with many trained architectures. Whenever a new architecture is trained, we initialize the architecture by the trained architectures in the pool. In this way, the number of training epochs for the new architecture can be reduced without harming the reliability of performance estimation Figure~\ref{fig:sub.pretrain3}. 

\section{Preliminary Observation}
\label{section3}

RL-based NAS \hs{generally} consumes \hs{quite expensive} computing resources. MNAS \cite{MnasNet} needs to train 8,000 models for training its RL agent until convergence, which costs 20,000 GPU hours on V100. Each architecture sample trained for one NAS process would not be used again. However, state-of-the-art differentiable-based NAS \cite{DARTS, xie2018snas, PDARTS, wu2019fbnet} demonstrated that, with various weight-sharing techniques, the NAS algorithms can be \hs{significantly} accelerated. In this section, we will show that the knowledge of previous searched processes can be reused,
which can accelerate the NAS processes.

\subsection{Architecture knowledge can be transferred} \label{section31}


Optimal architectures for different tasks have common architecture knowledge.
\hs{It can be observed in many applications that a good network architecture in one task tends to generalize to work well on other tasks.}
\hs{An illustrative example} of the observation is \hs{adopting the pre-trained} ImageNet \cite{imagenet} models \hs{as the backbone networks for} object detection \cite{coco}, semantic segmentation \cite{lin2017refinenet}, face recognition \cite{arcface}, etc. In NAS, however, this assumption needs to be carefully \hs{verified} as there exist a huge search space of \hs{network architectures} 
Here, We statistically verify \hs{whether this observation also holds for NAS}.

\begin{figure*}[!t]
\centering
\includegraphics[width=1\linewidth]{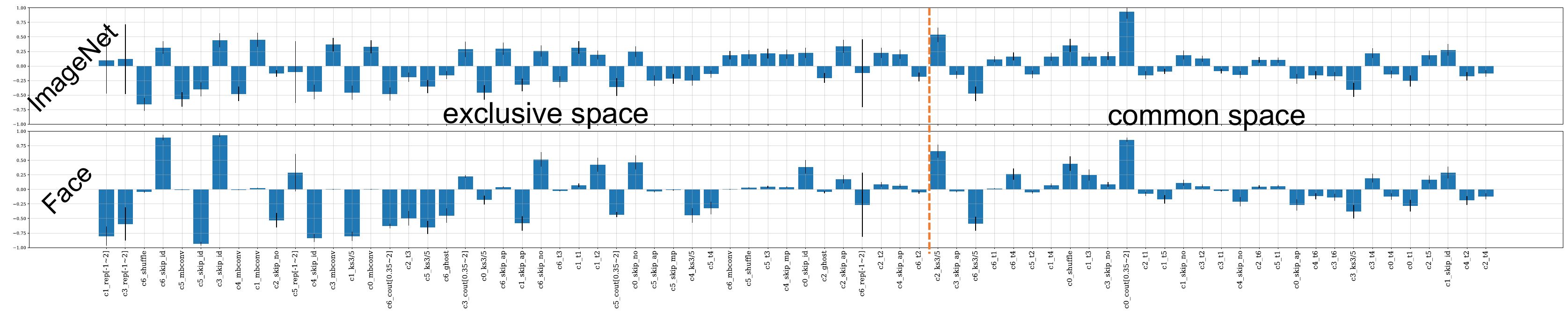}
\caption{Expectation of each operator of optimal models of face and ImageNet architecture search processes. Calculated by the 100 optimal models of face and ImageNet architecture search processes and sorted by the significance of the difference.}
\vspace{-2pt}
\label{fig:distance}
\end{figure*}

\begin{figure*}[t!]
    \centering
    \begin{subfigure}[b]{0.3\textwidth}
        \centering
        \includegraphics[width=\textwidth]{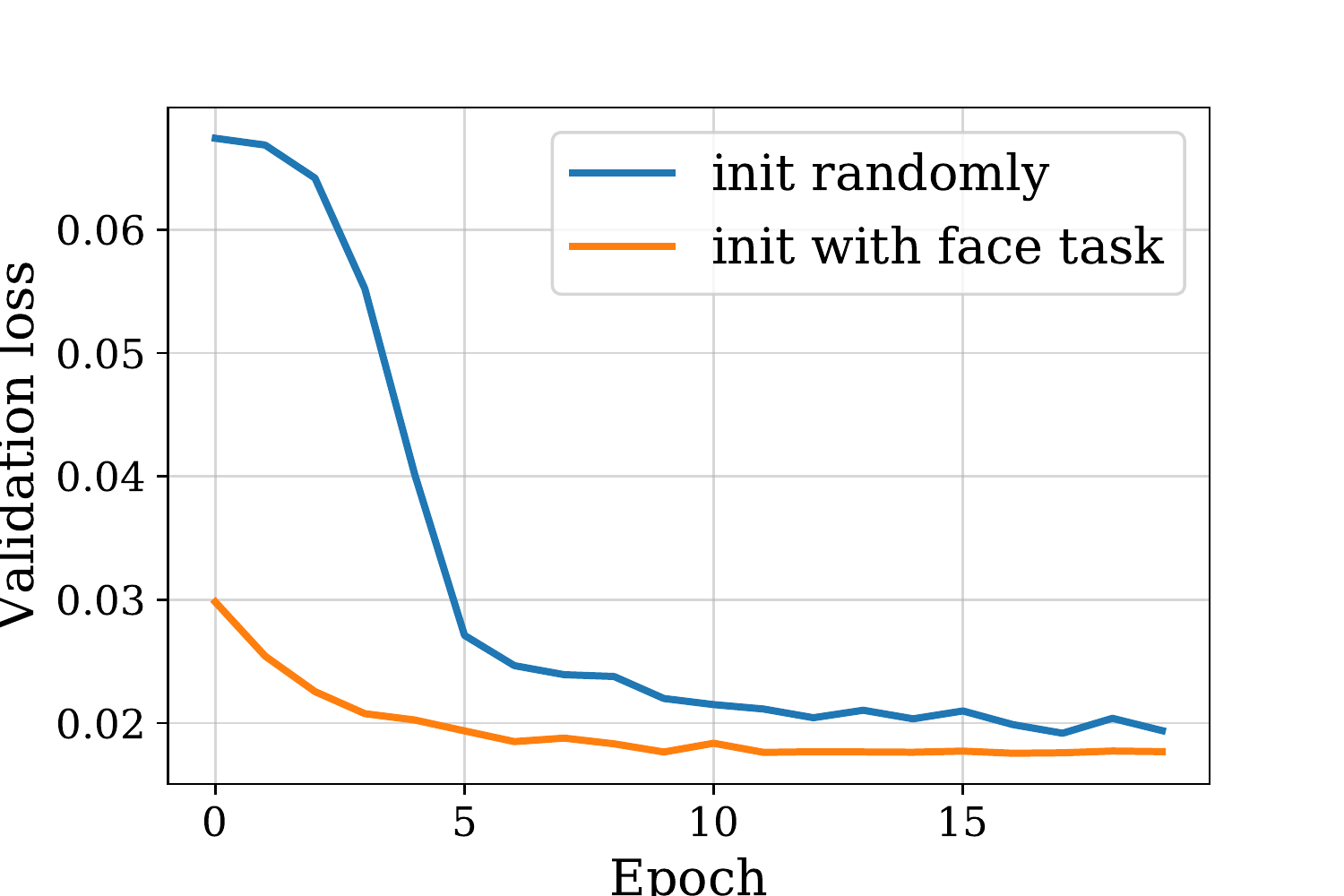}
        \caption{}
        \label{fig:sub.pretrain1}
    \end{subfigure}
    \centering
    \begin{subfigure}[b]{0.3\textwidth}
        \centering
        \includegraphics[width=\textwidth]{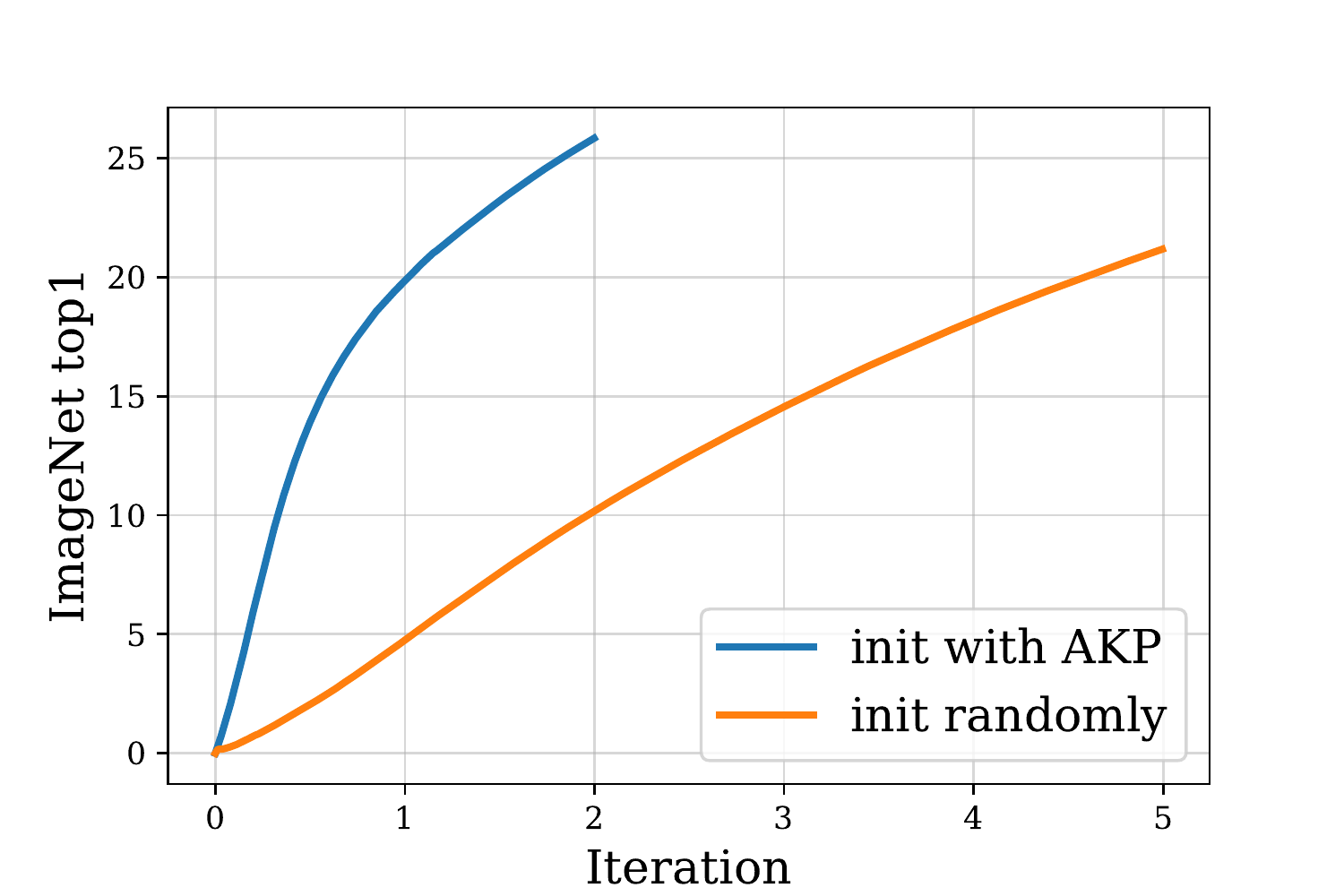}
        \caption{}
        \label{fig:sub.pretrain2}
    \end{subfigure}
    \begin{subfigure}[b]{0.3\textwidth}
        \centering
        \includegraphics[width=\textwidth]{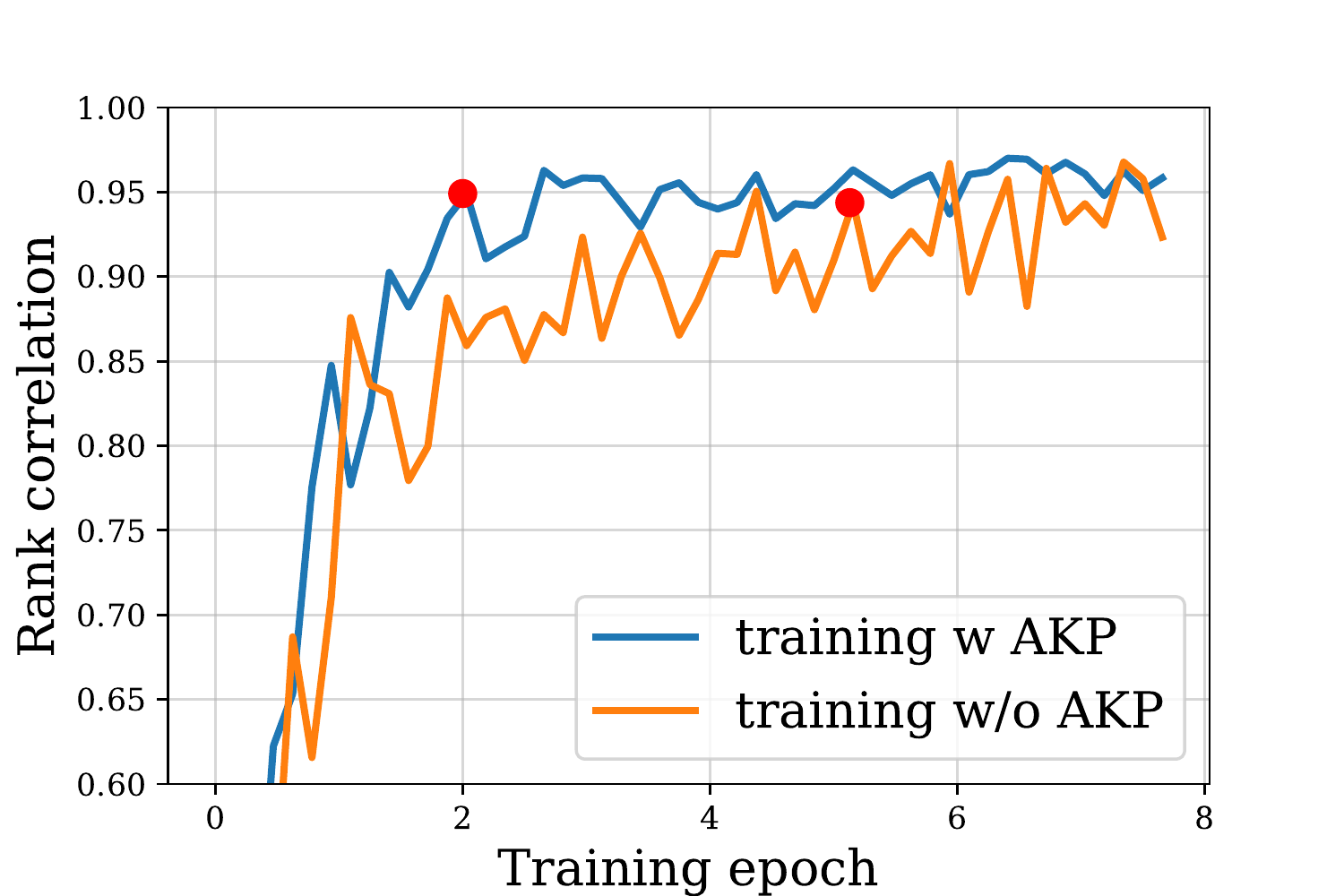}
        \caption{}
        \label{fig:sub.pretrain3}
    \end{subfigure}
\caption{On the left, the value function pretrained on face recognition tasks converges much faster. On the right, Spearman rank-order correlation \cite{spearman} along the training process of random initialization and block-level initialization.} 
\label{fig:pretrain}
\end{figure*}

In Figure~\ref{fig:distance}, we sample 100 optimal architectures of one face recognition search process and one ImageNet classification search process, respectively. For each architecture, we firstly expand its tokens to one-hot representation following \cite{liuyu}. After that, we can compare statistical divergence between the architecture family of face and ImageNet search processes. The results are shown in Figure~\ref{fig:distance}. Similar conclusions can be obtained that the operators can be divided into two categories, one with large differences and the other with small differences.

Many previous works \cite{pnas, kokiopoulou2019fast, luo2018neural, wen2019neural, seminas} use a predictor to predict an architecture's performance to speed up the NAS process. However, as the predictor requires thousands of samples to train, they usually \hs{evolves} in a progressive \cite{pnas} or semi-supervised manner \cite{seminas}. Inspired by the interesting observation above, we implement it in a unified way where different search processes' samples are used together to train a unified value network to map each architecture's one-hot representation \cite{liuyu} to its performance.
When running a new search experiment, we just use directly the unified network trained by the old data and keep updating it in the new task during the search process, which speeds up the convergence of the value network. As shown in Figure~\ref{fig:sub.pretrain1}, when transferring a value network trained on ImageNet to face recognition task, the network converges much faster.

\subsection{Parameter knowledge can be transferred} \label{section32}
Initializing the network by ImageNet pre-trained models and training the model on other tasks has generally been a standard way as it can speed up the convergence process. However, pretraining has been ignored in NAS as it may break the rank orders of different models. In our experiments, we observe that the parameter knowledge can help us to obtain the accurate rank correlations faster than training from scratch. Besides, this property holds regardless of the data distribution. We randomly sample 50 models and train them on ImageNet in two ways: from scratch or by initializing with \textit{parameter knowledge} from face recognition models. Then, we compare the rank order of validation set performance with the actual rank (i.e. fully trained rank) along the training process. As shown in Figures~\ref{fig:sub.pretrain2} and \ref{fig:sub.pretrain3}, with \textit{parameter knowledge} from face \hs{recognition models}, we can obtain more accurate rank in fewer epochs.

\begin{figure*}[t!]
\centering
\centerline{\includegraphics[width=\textwidth]{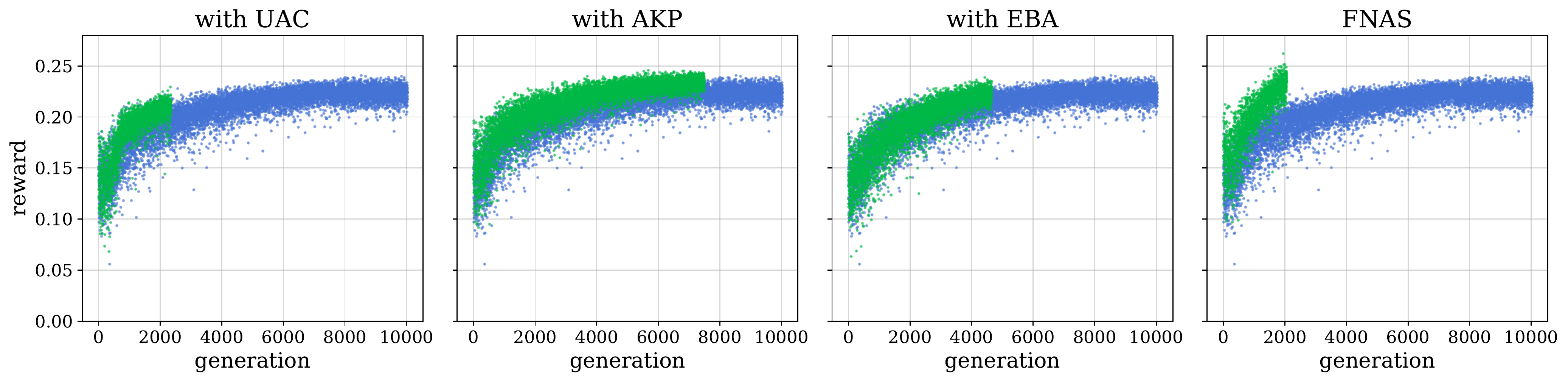}}
\caption{Reward along sample generation between FNAS and MNAS. Blue dots are the searching result of MNAS, while green dots are the results of FNAS.}
\label{fig.com}
\end{figure*}

\section{Uncertainty-Aware Neural Architecture Search}
\label{section4}
In this section, we introduce how we utilize the observations above to design three core modules to inherit common architecture knowledge and parameter knowledge from \hs{other tasks}.

\subsection{Architecture search with Reinforcement Learning} \label{section41}
Following \cite{MnasNet, NASNet}, we use Proximal Policy Optimization (PPO) \cite{ppo} to find our Pareto optimal solutions for our multi-objective search problem. Concretely, we follow the same idea as \cite{MnasNet, NASNet} and map each sample architecture in the search space to a list of tokens. These tokens are determined by a sequence of actions $a_{1:T}$ from the RL agent with policy $\pi_\theta$. Our goal is to maximize the expected reward 

\begin{equation} \label{eq1}
    \mathbb{J} = \mathbb{E}_{P_{(a_{1:T};m)}}R(m)
\end{equation}

where $m$ is a sampled model determined by action $a_{1:T}$, and $R(m)$ is the reward of $m$. We use the same definition of $R(m)$ of \cite{MnasNet} for fair comparison 
\begin{equation} \label{eq2}
    R(m) = {{ACC(m)} \times {\left[{\frac{LAT(m)}{T}}\right]^\alpha}},
\end{equation}
where $ACC(m)$ is the accuracy on the proxy task, $LAT(m)$ is the latency on target hardware, $T$ is the target latency, and $\alpha$ is the weight factor.

Following \cite{MnasNet}, we use a well known sample-eval-update loop to update the policy $\pi_\theta$. At each iteration, $\pi_\theta$ firstly generates a batch of samples by predicting a sequence of tokens with its LSTM. For each sample $m$, we train it on the proxy task to obtain $ACC(m)$ and run it on target hardware to obtain $LAT(m)$. $R(m)$ is calculated with Eq. (\ref{eq2}). We then update the policy $\pi_\theta$ to maximize the expected reward (Eq. (\ref{eq1})) using PPO.

\subsection{Uncertainty-Aware Critic in Proximal Policy Optimization (PPO)}
\label{section42}

The value function is a common module and is widely used in RL algorithms \hs{such as} PPO but rarely used in traditional NAS. Usually, training a value function requires a large number of samples (e.g., million-level steps in Atari {environment} trained by ray\footnote{https://github.com/ray-project/rl-experiments}), which is unbearable for NAS as it \hs{is equivalent to training} thousands of models needed to be trained  and it's expensive. In our algorithm, we propose the Uncertainty-Aware Critic (UAC) to deal with this issue, which is inspired by \hs{our observations as mentioned in} Section~\ref{section3}. 

Given an architecture $m$ sampled from \hs{the} search space, a value network $V$ is utilized to predict the reward $V(m)$ of this sample, while $R(m)$ is the \hs{actual} reward of it. The loss function to update $V$ is \hs{formulated as}
\begin{equation}
    L_V = |V(m)-R(m)|
\end{equation}

Besides, an uncertainty network $U$ is utilized to predict the uncertainty $U(m)$ of this \hs{sampled architecture $m$}, which is used to learn discriminately whether a sample is in the distribution of learned samples. The loss function to supervise $U$ is formulated as

\begin{equation}
    L_U = |U(m)-L_V|
\end{equation}

If $U(m)$ is greater than a threshold, the sample may locate in an untrusted region, which \hs{indicates that the sampled architecture $m$} has not been \hs{effectively} learned by the value network. As a result, it would be trained on a proxy task from scratch to get its reward $R(m)$. Otherwise, \hs{it can be assumed that} the prediction $V(m)$ is accurate, and $V(m)$ would be regarded as $R(m)$ to update the RL agent. The threshold is set to ensure 2 times speedup while avoiding the risk of over-fitting. The whole process is illustrated in Figure~\ref{fig.pipeline}. 

Samples \hs{that} need to be trained from scratch are \hs{named} as untrusted samples, while \hs{the remaining ones whose reward comes from $V$} are named as trusted samples. With more untrusted samples \hs{obtaining} their rewards along the search process, the value network becomes more accurate, thus the uncertainty \hs{predicted by} $U$ \hs{gradually} decreases. Considering an extreme case, where each sample in a batch \hs{obtains} a reward with low uncertainty and is classified as trusted sample, the agent \hs{trained with} these samples is \hs{likely to overfit}, which is not conducive to the exploration of the RL agent and would lead to \hs{inferior} performance.
In our implementation, we use the following constraint to balance the exploration and the exploitation of the RL agent to speed up its convergence without over-fitting.

\begin{itemize}[leftmargin=2em]
    
    \item Constraint: In each batch, when the number of trusted samples is greater than 50\% of the batch size, the extra trusted samples \hs{would} be thrown away and the \hs{architectures would be} resampled until enough untrusted samples are \hs{obtained} to fill the batch.
\end{itemize}
With \hs{the above} constraint, the algorithm can get a decent performance \hs{with} accelerated search, and the result is shown in Figure~\ref{fig.com} (with UAC).

\begin{wrapfigure}{r}{7.5cm}
\centering
\centerline{\includegraphics[width=1\columnwidth]{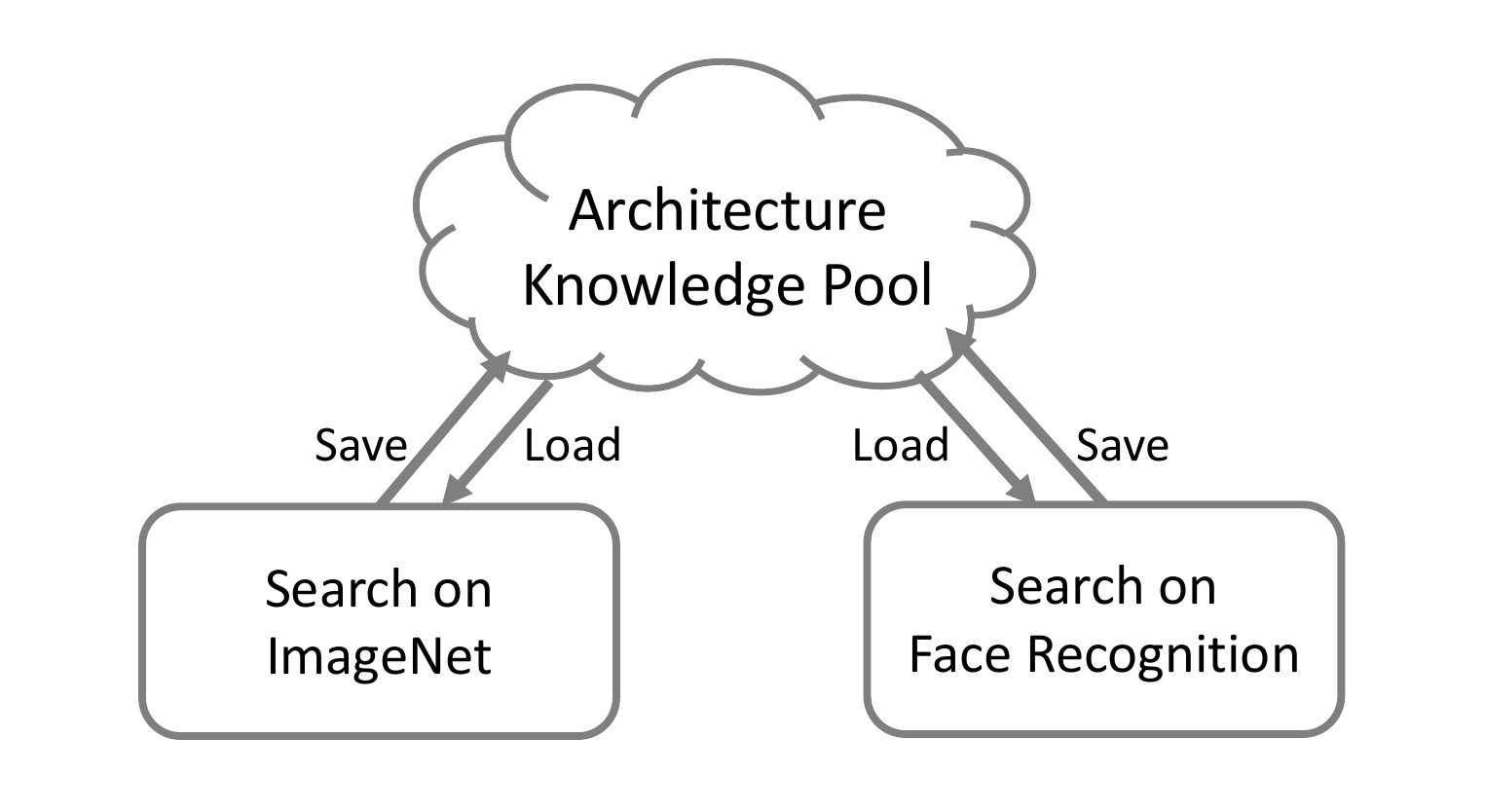}}
\caption{Different tasks share the same global knowledge pool.} 
\label{fig:cloud}
\end{wrapfigure}

\subsection{Uncertainty-Aware Architecture Knowledge Pool} \label{section43}

\textit{Parameter knowledge} can be transferred among different tasks to speed up the convergence of the training process of the sampled architectures as shown in Section~\ref{section32}. However, traditional pretrain is not feasible in NAS, as there are thousands of different architectures in the search space and we can not afford to pretrain each architecture on a different task. To address this problem, we propose to initialize each architecture in a factorized way and use a fuzzy matching algorithm to guarantee the hit ratio, which is defined as the division between the number of matched blocks and total queried blocks.

Following \cite{MnasNet}, we define an architecture as a combination of $n$ blocks $\{b_1, b_2, \dots, b_n\}$. For any two architectures $m_i$ and $m_j$, although generally, their structures might be quite different, some of their blocks are similar to each other, (e.g., $b_2$ of $m_i$ == $b_2$ of $m_j$), thus the weights of these parts could be shared. So we build a Architecture Knowledge Pool (AKP) to store all the previously trained models' \hs{blocks} in a key-value \hs{table}, where the key is the expand embedding \cite{liuyu} of each block and the value is the \textit{Parameter knowledge} of the block.

Recent research has found that fairness in weight sharing has great influence on the final performance of searched architecture \cite{fairnas}. So we apply the following two strategies to solve the problem of fairness.

\begin{itemize}[leftmargin=2em]
    \item The checkpoints stored in the AKP are trained with equal iterations.
    \item For each block query, the proposed uncertainty function is used to ensure that the match ratio reaches more than 99\%, which means less than 1\% blocks have been unfairly initialized.
\end{itemize}

Given a query block $b_i$, we calculate the cosine similarity of the expanded embedding as in Section~\ref{section31} between $b_i$ and each element in AKP. The block with the highest similarity \hs{would be retrieved} to initialize $b_i$. We show the overall process in Figure~\ref{fig:cloud}. \hs{Our experiments shows that} using AKP \hs{created from multiple tasks} can speed up the search process by 2$\times$ (Figure~\ref{fig:sub.pretrain3}), as shown in Figure~\ref{fig.com} (with AKP).

\section{Architecture Experience Buffer (3AEB)} \label{section5}

In a general RL task, there are a lot of discussions about sample reuses. However, in RL-based NAS, \hs{improving sampling} efficiency is rarely \hs{investigated}. In our algorithm, we propose \hs{an} architecture experience buffer to store the sampled \hs{architectures} in the form of architecture-performance pairs, and for each iteration in the future, the stored samples may be used again to update the RL agent to speed up its convergence. We call the samples stored in the experience buffer as exploited sample and the newly generated samples as exploring samples. Different from the traditional RL works, the proposed experience buffer has the following features:

\begin{itemize}[leftmargin=2em]
    \item The buffer size is relatively small (usually 10 in our experiments). As the convergence of the RL agent is much faster than RL tasks, if the buffer size is set too large, the agent will focus on exploited samples and the convergence speed would be slow.
    
    \item In each batch, both exploited samples and exploring samples would be selected. To prevent the RL updating from biasing to the exploited samples, the percentage of the exploited samples in one batch is constrained to no more than 50\%.
\end{itemize}

Some recent works~\cite{priority} suggested that the samples \hs{of different properties} should be selected in the buffer. We \hs{follow the strategy by choosing samples with different reward values}. Then, we sample from the buffer and reweight \hs{the} samples \hs{according to} their priorities \hs{as defined below}.
For each sample $\{s_1, s_2, \dots, s_n\}$ with their rewards $R \{r_1, r_2 \dots, r_n\}$ in AEB, \hs{their} priority \hs{scores are} defined as
\begin{equation}
    P_i = \frac{exp(r_i)}{\sum_{j}^{ }exp(r_j)}
\end{equation}

Following~\cite{priority}, each sample \hs{would also} be reweighted by \hs{their} importance sampling \hs{weights}. The reweighted \hs{priority scores} $S$ can be written \hs{as} $S_i = (N \cdot P_i)^{-\beta}$, where $N$ is \hs{the} buffer size and $\beta$ is the annealing \hs{coefficient} and \hs{would be increased} from 0 to 1 as the search proceeds. And the result is shown in Figure~\ref{fig.com} (with AEB).

\begin{table*}[]
\caption{Performance Results on ImageNet Classification. FNAS-Image$\times$1.3 means scale up FNAS-Image for 1.3$\times$ along width.}
\centering
\begin{tabular}{c|c|c|cc|c}
\midrule
models                  & Type                            & Mult-Adds         & Top1 Acc. (\%) & Top5 Acc. (\%) & \begin{tabular}[c]{@{}c@{}}Search Cost \\ (GPU Hours)\end{tabular} \\ \midrule
MBv2                    & Manual                          & 300M          & 72             & 91             & 0                                                                 \\ \midrule
ProxylessNAS            & \multirow{4}{*}{Share-weight} & 320M          & 74.6           & 92.2           & 200                                                               \\
DARTS                   &                                 & 574M          & 73.3           & 91.3           & 96                                                                \\
FairNAS                 &                                 & 388M          & 75.3           & 92.4           & 288                                                               \\
Once-For-All            &                                 & 327M          & 75.3           & 92.6           & 1200                                                              \\ \midrule
AmoebaNet               & Evolutionary                    & 555M          & 74.5           & 92             & 75,600                                                            \\ \midrule
MNAS                    & \multirow{6}{*}{RL-based}       & 315M          & 75.2           & 92.5           & 20,000                                                            \\
NASNet                  &                                 & 564M          & 74             & 91.6           & 43,200                                                            \\
MBv3                    &                                 & 219M          & 75.2           & 91             & -                                                                 \\
EfficientNetB0          &                                 & 390M          & 76.3           & 93.2           & -                                                                 \\
\textbf{FNAS-Image}     &                                 & \textbf{225M} & \textbf{75.5}  & \textbf{92.6}  & \textbf{2000}                                                     \\
\textbf{FNAS-Image$\times$1.3} &                                 & \textbf{392M} & \textbf{77.2}  & \textbf{93.5}  & \textbf{2000}                                                     \\ \midrule
\end{tabular}
\label{tab:mainresults}
\end{table*}


\section{Fast Neural Architecture Search (\modelnameshort{}) on vision tasks} \label{section_exp}
In this section, we conduct different experiments on both ImageNet and million-level face recognition tasks to verify the effectiveness of FNAS. The details and results are as follows: 

\subsection{Implementation details} \label{exp_detail}

Following the standard searching algorithm as NASNet \cite{NASNet}, MNAS \cite{MnasNet} and AKD \cite{liuyu}, we use an RNN-based agent optimized by PPO algorithm \cite{ppo}. The RL agent is implemented with a one-layer LSTM \cite{lstm} with 100 hidden units at each layer.
The $V$ and $U$ of UAC are implemented with four-layer MLP with 200 hidden units at each layer and PReLU \cite{prelu} nonlinearity. 
 
For ImageNet experiments, we sample 50K images from the training set to form the mini-val set and use the rest as the mini-training set. In each experiment, 8K models are sampled to update the RL agent. Note that when equipped with UAC or AEB, not all samples need to be activated, as many samples' rewards are directly returned from these two modules. For face experiments, we use MS1M \cite{ms1m} as the mini-training set, LFW \cite{lfw} as the mini-val set. The final performance is evaluated on MegaFace \cite{megaface}. 

\begin{table*}
\begin{floatrow}
\begin{minipage}{0.1\linewidth}
\centering
\renewcommand\tabcolsep{4pt}
\ttabbox{
 \caption{Performance on MegaFace.}
 \label{tab:faceresult}
}{%
 \begin{tabular}{c|c|cc|c}
\midrule
\multirow{2}{*}{model} & \multirow{2}{*}{Mult-Adds} & \multicolumn{2}{c|}{Distractor num} & \multirow{2}*{\shortstack{Search Cost\\ (GPU Hours)}} \\
                      &                        & 1e5              & 1e6              &                                                                  \\ \midrule
MBv2                   & 300M                   & 92.75            & 88.71            & 0                                                                \\
ShuffleNet             & 295M                   & 94.15            & 90.46            & 0                                                                \\
MNAS                   & 313M                   & 93.41            & 89.47            & 20,000                                                             \\
MBv3                   & 218M                   & 94.15            & 90.64            & -                                                                \\
\textbf{FNAS-Face}          & \textbf{227M}          & \textbf{95.45}   & \textbf{92.63}   & \textbf{2000}                                                      \\ \midrule
 \end{tabular}
}
\end{minipage}

\begin{minipage}{0.1\linewidth}
\centering
\renewcommand\tabcolsep{4pt}
\ttabbox{
 \caption{Performance on COCO.}
 \label{tab:detection}
}{%
 \begin{tabular}{c|c|c}
\midrule
models                   & Mult-Adds    & mAP      \\ \midrule
MNAS                     & 13.4 G  & 27.68     \\
MBv2$\times$1.0          & 13.3 G  & 29.79  \\
MBv3$\times$0.75         & 5.852 G  & 29.25  \\
MBv3$\times$1.0          & 9.060 G  & 30.02  \\ \midrule
\textbf{FNAS}        & \textbf{8.021 G}  & \textbf{30.44}     \\ \midrule
 \end{tabular}
}
\end{minipage}
\end{floatrow}
\end{table*}

\subsection{Proxyless FNAS on ImageNet}

Just as MNAS \cite{MnasNet} has done, we also use a multi-objective reward to directly search on ImageNet. After the search process, we retrain the top 10 models with the largest reward near the target Mult-Adds from scratch to verify the search results. In Table~\ref{tab:mainresults}, we get a relatively higher result than the current SOTA network MBv3 \cite{mbv3}. Note that the model we search does not go through the pruning operation NetAdapt \cite{netadapt}, which can reduce 10\%$\sim$15\% computation and keep performance nearly unchanged. Compared with EfficientNetB0 \cite{efficientnet}, FNAS improves top 1 accuracy by 1 point under comparable computation budget.  And still, there is nearly 10$\times$ of acceleration in the entire search process compared to MNAS \cite{MnasNet} or MBv3 \cite{mbv3}.

\subsection{Proxyless FNAS on fine-grained facial recognition}

Besides verifying the performance of FNAS on ImageNet, we also test it on the fine-grained facial recognition task. As can be seen in Table~\ref{tab:faceresult}, compared with MBv3, verification accuracy improves 2 points in comparable Mult-Adds under 1e6 distractors. When compared with MBv2, FNAS improves verification accuracy for nearly 4 points with 24\% Mult-Adds reduction. 
The result shows: 1) FNAS has an obvious acceleration effect on different tasks and 2) the importance of searching directly on the target task.

\subsection{Transferability on object detection}

We combine the model found on ImageNet in Table~\ref{tab:mainresults} with the latest pipeline of detection to verify its generalization. Table~\ref{tab:detection} shows the performance of the model on COCO \cite{coco}. It can be seen that compared to MBv3, there is a significant improvement with our searched model.


\begin{table*}[]
\tiny
\setlength{\abovecaptionskip}{0pt}
\caption{The effectiveness of the three proposed modules, MBv2$\times$0.38 means scale up MBv2 for 0.38$\times$ along width}
\vspace{0mm}
\resizebox{\textwidth}{!}{
\begin{tabular}{c|ccc|cc|ccc}
\midrule
Models                & AKP                            & UAC                            & AEB                            & Mult-Adds      & Top1 Acc. (\%) & \begin{tabular}[c]{@{}c@{}}Activated\\ Samples\end{tabular} & \begin{tabular}[c]{@{}c@{}}Search\\ Epoches\end{tabular} & \begin{tabular}[c]{@{}c@{}}Search Cost\\ (GPU Hours)\end{tabular} \\ \midrule
MBv2$\times$0.38       & \multicolumn{3}{c|}{}                                                                            & 81M          & 62.65          & 0                                                           & 0                                                        & 0                                                                 \\ \midrule
\multirow{2}{*}{MNAS} & \multicolumn{3}{c|}{\multirow{2}{*}{}}                                                           & 72M          & 64.23          & 8,000                                                       & 1                                                        & 4,000                                                             \\  
                      & \multicolumn{3}{c|}{}                                                                            & 74M          & 65.19          & 10,000                                                      & 4                                                        & 20,000                                                            \\ \midrule
\multirow{5}{*}{FNAS} & \cmark          &                                &                                & 75M          & 64.97          & 8,000                                                       & 1                                                        & 4,000                                                             \\ 
                      & \cmark          &                                &                                & 76M          & 65.22          & 8,000                                                       & 2                                                        & 10,000                                                            \\ 
                      &                                & \cmark          &                                & 72M          & 64.28          & 2,300                                                       & 1                                                        & 1,150                                                             \\ 
                      &                                &                                & \cmark          & 72M          & 64.44          & 4,500                                                       & 1                                                        & 2,250                                                             \\ 
                      & \textbf{\cmark} & \textbf{\cmark} & \textbf{\cmark} & \textbf{85M} & \textbf{66.25} & \textbf{2,000}                                              & \textbf{1}                                               & \textbf{1,000}                                                    \\ \midrule
\end{tabular}}
\label{tab:ablation1}
\end{table*}

\begin{table*}[]
\tiny
\setlength{\abovecaptionskip}{0pt}
\vspace{0mm}
\caption{Transferability of UAC and AKP}
\resizebox{\textwidth}{!}{
\begin{tabular}{c|c|cc|ccc}
\midrule
models   & UAC or AKP & Mult-Adds & Top1 Acc. (\%) & \begin{tabular}[c]{@{}c@{}}Activated \\ Samples\end{tabular} & \begin{tabular}[c]{@{}c@{}}Search \\ Epoches\end{tabular} & \begin{tabular}[c]{@{}c@{}}Search Cost\\ (GPU Hours)\end{tabular} \\ \midrule
MNAS & \xmark                    & 181M & 73.25 & 8,000 & 4 & 16,000 \\ \midrule
\textbf{FNAS}     & UAC init with face exp & \textbf{153M}    & \textbf{73.91} & \textbf{2,000} & \textbf{4} & \textbf{4,000} \\ \midrule
MNAS & \xmark                & 285M & 74.62 & 8,000 & 4 & 16,000  \\ \midrule
\textbf{FNAS}     & AKP init with face exp & \textbf{292M} & \textbf{75.22} & \textbf{4,000}  & \textbf{4}  & \textbf{8,000}  \\ \midrule
\end{tabular}}
\vspace{-5mm}
\label{tab:ablation2}
\end{table*}

\section{Ablation study}

\subsection{The effectiveness of the three proposed modules.}

In this section, the effectiveness of Uncertainty-Aware Critic (UAC), Architecture Knowledge Pool (AKP), Architecture Experience Buffer (AEB) is verified when they are used alone or combined. Details are shown in Table~\ref{tab:ablation1}. Three conclusions can be observed:
1. Sampling with AKP initialization gets real rank faster;
2. Fewer samples are required when NAS is equipped with UAC and AEB; and 
3. 10$\times$ speedup can be achieved when NAS is equipped with AKP, UAC, and AEB.

\subsection{The transferability of the proposed modules.}

In Section~\ref{section3}, we mentioned that knowledge between NAS processes is transferable, which is also verified in the experiment. We use the UAC trained on the face as a pre-trained model and then transfer it to the ImageNet architecture search process. In the absence of 3/4 of activated samples, the optimal model surpasses baseline by 0.67\% with fewer Mult-Adds, showing in Table \ref{tab:ablation2}. In addition, we use AKP with the checkpoints from face architecture search process and then search on ImageNet. In the absence of 1/2 activated samples, performance increases by 0.6\%.

\section{Conclusion}
\label{sec:con}

This paper proposes three modules (UAC, AKP, AEB) to speed up the entire running process of RL-based NAS, which consumes large amounts of computing power before. With these modules, fewer samples and less training computing resources are needed, making the overall search process 10$\times$ faster. We also show the effectiveness of applying those modules on different tasks such as ImageNet, face recognition, and object detection. More importantly, the transferability of UAC and AKP is being tested by our observation and experiments, which will guide us in tapping the knowledge of the NAS process.

{\small
\bibliographystyle{ieee_fullname}
\bibliography{neurips_2021}
}

\end{document}